\documentclass[11pt, a4paper]{article}
\usepackage[hyperref]{emnlp2018}
\usepackage{times}
\usepackage{latexsym}

\usepackage{url}
\usepackage{changepage}
\usepackage{pifont}
\usepackage{caption}
\usepackage{subfigure}
\usepackage{amsmath, bm}
\usepackage{verbatim}
\usepackage{dcolumn}
\usepackage{polynom}
\usepackage{color}
\usepackage{mdwlist}
\usepackage{hyperref}

\usepackage{diagbox}
\usepackage{multirow}
\usepackage{booktabs}  
\usepackage{threeparttable} 
\usepackage{pgfplots}
\usepackage{setspace}

\usepackage{lipsum}
\usepackage{enumitem,calc}
\usepackage[normalem]{ulem}

\usepackage{array}

\newlist{modelStep}{description*}{3}
\newlist{case}{description*}{1}

\aclfinalcopy 

\usepackage{CJKutf8}

\title{A Syntactically Constrained Bidirectional-Asynchronous Approach for Emotional Conversation Generation}

\author{Jingyuan Li, Xiao Sun\thanks{\ \ The corresponding author of this paper.}\\
  Anhui Province Key Laboratory of Affective Computing\\
   and Advanced Intelligent Machine\\
  Hefei University of Technology, Hefei, China\\
  {\tt lijingyuan@mail.hfut.edu.cn, sunx@hfut.edu.cn}
  }

\date{}

\begin{document}
\maketitle
\begin{abstract}
Traditional neural language models tend to generate generic replies with poor logic and no emotion.
In this paper, a syntactically constrained bidirectional-asynchronous approach for emotional conversation generation (E-SCBA) is proposed to address this issue.
In our model, pre-generated emotion keywords and topic keywords are asynchronously introduced into the process of decoding.
It is much different from most existing methods which generate replies from the first word to the last.
Through experiments, the results indicate that our approach not only improves the diversity of replies, but gains a boost on both logic and emotion compared with baselines.
\end{abstract}

\section{Introduction}

In recent years, as artificial intelligence has developed rapidly, researchers are pursuing technologies with greater similarities to human intelligence.
As a subjective factor, emotion performs an elemental difference between humans and machines. 
In other words, machines that could understand emotion would be more responsive to human needs. 
For example, in education, positive emotions improve students' learning efficiency \cite{Kort2002An}. 
In healthcare, mood prediction can be used in mental health counseling to help anticipate and prevent suicide or depression \cite{taylor2017personalized, jaques2017multimodal}.
To make machine more intelligent, we must resolve the conundrum of emotional interactions.

There are tons of researches about conversation, an important channel for communication between humans.
And lots of work has recently been carried out in open-domain conversation devoted to generating meaningful replies \cite{vinyals2015neural, li2016diversity, serban2016building}. 
Unfortunately, the factors considered in these methods only concerns topic, like \cite{xing2017topic}, where they failed to take emotion into account.
Unlike the former, the work in \cite{zhou2017emotional} first addressed the emotional factor in large-scale conversation generation, and it showed that emotional replies obtain superior performances compared to the baselines that did not consider emotion.
However, two defects still manifest themselves in the aforementioned models.
{\bfseries First}, all methods above only adopted a single factor (i.e., topic or emotion), because of which the bias of information can not comprehensively summarize the human conversations to achieve favorable results.
{\bfseries Second}, the way that generates replies from the first word to the last can lead to a decline in diversity, limited by the high-frequency generic words in the beginning (e.g., I and you), as argued in \cite{mou2016sequence}. 

The deficiencies above inspire us to introduce a new approach called E-SCBA, studying both emotion and topic. Three main contributions are presented in this paper:
(1) It conducts a study of compound information, which constitutes the syntactic constraint in the conversation generation.
(2) Different from the work in \cite{mou2016sequence}, a bidirectional-asynchronous decoder with multi-stage strategy is proposed to utilize the syntactic constraint. It ensures the unobstructed communication between different information and allows a fine-grained control of the reply to address the problem of fluency and grammaticality as argued in \cite{ghosh2017affect, zhou2017emotional}.
(3) Our experiments show that E-SCBA work better on emotion, logic and diversity than the general seq2seq and other models that consider only a single factor during the generation.

\section{Model}
\label{sec:length}

\subsection{Overview}
\label{sec:overview}

\begin{figure*}[t]
	\centering
	\begin{center}
		\includegraphics[width=1.0\linewidth]{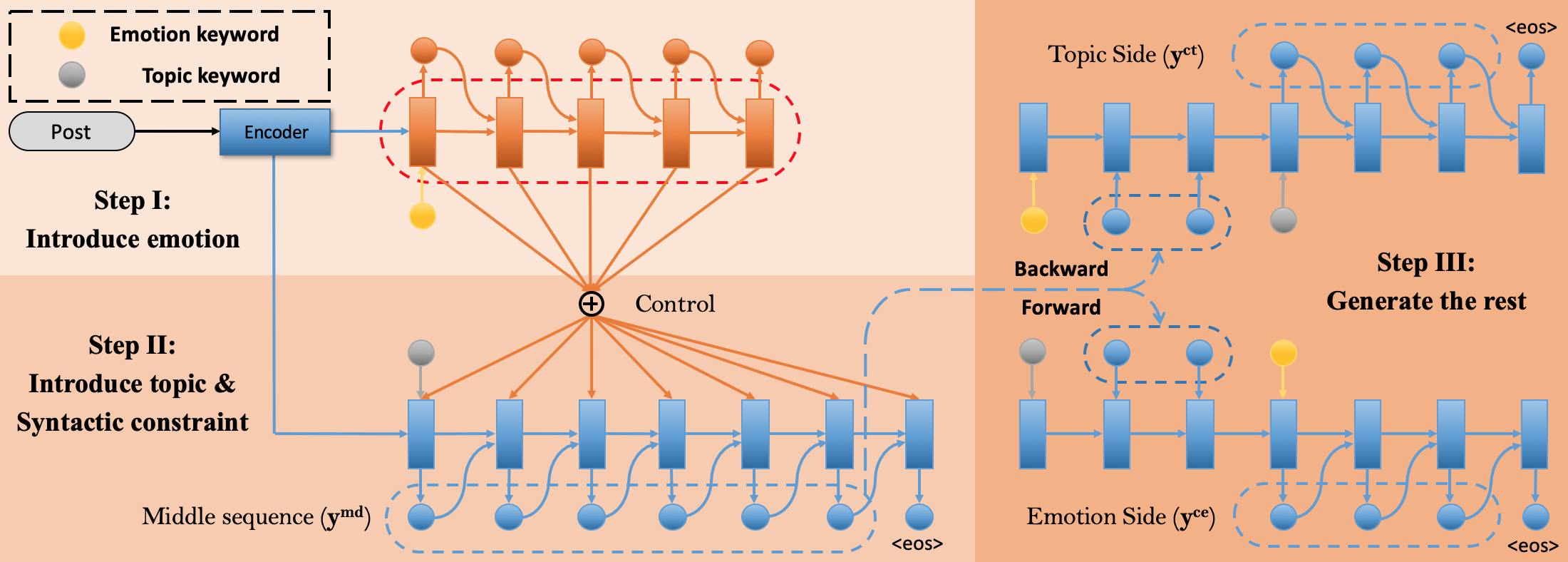}
	\caption{The process of generating replies in the test. The middle part of the reply is generated in Steps I and II, and the remaining two sides are generated in Step III. The RNN networks used in the decoder do not share the parameters with each other.}
	\label{fig:Decoder}
	\end{center}
\end{figure*}

The whole process of emotional conversation generation consists of the following three steps:
\begin{description}
\item[\emph{Step \uppercase\expandafter{\romannumeral1}}:]Given a post, we first use two networks combined with category embeddings to respectively predict emotion keyword and topic keyword that should appear in the final reply (see Section \ref{sec:wordPredictor}).
\item[\emph{Step \uppercase\expandafter{\romannumeral2}}:]After the prediction, a newly designed decoder is used to introduce both keywords into the content\footnote{Syntactic constraint starts to work here, and can be intuitively interpreted as relative positions of emotion words and topic words, as well as different combinations between them.}, as shown in Figure \ref{fig:Decoder}. It first produces a sequence of hidden states based on the emotion keyword (Step I), and then uses an emotional attention mechanism to affect the generation of middle sequence, which is based on the topic keyword (Step II). The remaining two sides are ultimately generated by the combination of middle part and keywords (Step III). A detailed description is given in Section \ref{sec:tripleDecoder}.
\item[\emph{Step \uppercase\expandafter{\romannumeral3}}:]Finally, a direction selector is used to arrange the generated reply in a logically correct order by selecting the better one from forward and backward forms of the reply generated in the last step (see Section \ref{sec:DS}).
\end{description}

In this work, we default that the replies contain at least one emotion keyword and one topic keyword, which are expected to appear in the dictionaries we used.

\subsection{Keyword Predictor}
\label{sec:wordPredictor}

The keywords to be selected are pre-stored in the prepared dictionaries.
The adopted emotion dictionary was proposed by \cite{Xu2008Constructing}, which contains 27,466 emotion words divided into 7 categories: \emph{Happy}, \emph{Like}, \emph{Surprise}, \emph{Sad}, \emph{Fear}, \emph{Angry} and \emph{Disgust}.
The adopted topic dictionary was obtained by the LDA model \cite{blei2003latent}, including 10 categories with 100 words for each category. And to avoid situations in which emotion and topic keywords are predicted to be the same word, all the overlapping words in these two dictionaries default to emotion keywords.

The prediction of emotion and topic keywords follows the similar path.
We first derive topic category and emotion category from the post with two classifiers separately. To be more specific, the pre-trained LDA model is used for the topic category inference. And the work in \cite{sun2018detecting} is applied for emotion. The concrete model is an emotion transfer network. Given a specific external stimuli (e.g., a sentence), the network produce an emotional response, which is specifically an emotion category in this work.
After this, combining the sum of hidden states ${\tilde{h}} = \sum_{i = 1}^{T} {h}_{i}$ from encoder and the category embeddings $\boldsymbol{k}$ = \{${k}^{et}$, ${k}^{tp}$\}, keywords are predicted as follows:
{\setlength\abovedisplayskip{1em}
\setlength\belowdisplayskip{1em}
\setlength\textwidth{1em}
\begin{align}
	p({w}^{k}_{et} | \boldsymbol{x}, {k}^{et}) &= softmax(\mathbf{{W}}^{w}_{et}[{\tilde{h}}; {k}^{et}])\\[0.6em]
	p({w}^{k}_{tp} | \boldsymbol{x}, {k}^{tp}) &= softmax(\mathbf{{W}}^{w}_{tp}[{\tilde{h}}; {k}^{tp}])
\end{align}
}where ${w}^{k}_{et}$ and ${w}^{k}_{tp}$ separately represent the emotion keyword and topic keyword that are expected to appear in the reply.

\subsection{Bidirectional-Asynchronous Decoder}
\label{sec:tripleDecoder}

Due to the decoder architecture shown in Figure \ref{fig:Decoder}, we suppose the reply in this section is $\boldsymbol{y} = (\boldsymbol{{y}}^{ct}, {w}^{k}_{tp}, \boldsymbol{{y}}^{md}, {w}^{k}_{et}, \boldsymbol{{y}}^{ce})$\footnote{For the training data in opposite direction, we reversed the target replies to meet the requirement.} 
where $\boldsymbol{{y}}^{md}$ is the middle part between two keywords and $\boldsymbol{{y}}^{ct}$, $\boldsymbol{{y}}^{ce}$ represent the remaining sides connected to the topic keyword and emotion keyword. The generation of middle part $\boldsymbol{{y}}^{md} = ({y}_{1}^{md}, ..., {y}_{K}^{md})$ can be described as follows:
{\setlength\abovedisplayskip{1em}
\setlength\belowdisplayskip{1em}
\begin{align}
        {c}_{j}^{et} &= f_{att}^{et}({s}_{j-1}^{tp}, \{{s}_{i}^{et}\}_{i = 1}^{K'}) \\
	p(\boldsymbol{{y}}^{md} |\boldsymbol{x}, \boldsymbol{w}^{k}) 
	&= \prod_{j = 1}^{K}p({y}_{j}^{md} | {y}_{j-1}^{md}, {s}_{j}^{tp}, {c}_{j}^{et})
\end{align}}where $\boldsymbol{w}^{k} = \ <{w}^{k}_{et}, {w}^{k}_{tp}>$ represents the set of keywords, ${s}_{i}^{et}$ and ${s}_{j}^{tp}$ separately represent the decoding state of the steps that introduce emotion keyword and topic keyword. ${c}_{j}^{et}$ is the emotional constrain unit at time ${j}$, computing by the emotion control function $f_{att}^{et}$ as follows:
{\setlength\abovedisplayskip{1em}
\setlength\belowdisplayskip{1em}
\begin{align}
	{c}_{j}^{et} &= \sum_{i = 1}^{K'} \alpha_{j, i}^{et} {s}_{i}^{et} \\
	\alpha_{j, i}^{et} &= \frac{\exp(e_{j, i}^{et})}{\sum_{t = 1}^{K'} \exp(e_{j, t}^{et})} \\
	e_{j, i}^{et} &= (\mathbf{v}^{md}_{\alpha})^{T}tanh(\mathbf{W}^{md}_{\alpha}{s}_{j-1}^{tp} + \mathbf{{U}}^{md}_{\alpha}{s}_{i}^{et})
\end{align}}where $e_{j, i}^{et}$ represents the impact scores of the emotion state ${s}_{i}^{et}$ on the topic state ${s}_{j-1}^{tp}$. 

After generating the middle part, we connect it with the keywords to form a new sequence. Two seq2seq models are used to encode the connected sequences and decode $\boldsymbol{{y}}^{ce} = ({y}_{1}^{ce}, ..., {y}_{M}^{ce})$ and $\boldsymbol{{y}}^{ct} = ({y}_{1}^{ct}, ..., {y}_{N}^{ct})$, as below:
{\setlength\abovedisplayskip{1em}
\setlength\belowdisplayskip{1em}
\begin{align}
	p(\boldsymbol{{y}}^{ce} | \boldsymbol{w}^{k}, \boldsymbol{{y}}^{md}) 
	&= p(\boldsymbol{{y}}^{ce} | [{w}^{k}_{tp}, \boldsymbol{{y}}^{md, f}, {w}^{k}_{et}])\\[0.6em]
	p(\boldsymbol{{y}}^{ct} | \boldsymbol{w}^{k}, \boldsymbol{{y}}^{md}) 
	&= p(\boldsymbol{{y}}^{ct} | [{w}^{k}_{et}, \boldsymbol{{y}}^{md, b}, {w}^{k}_{tp}])
\end{align}}where $\boldsymbol{{y}}^{md, f}$ and $\boldsymbol{{y}}^{md, b}$ are the forward and backward situations of the middle part, respectively.

\subsection{Direction Selector}
\label{sec:DS}

To make the samples meet the requirements of decoder, by default we place the topic keyword as the first keyword on the left and the emotion keyword on the right in training.
However, in real situations, the topic keyword does not always appear before the emotion keyword, where we must determine correct direction by the machine.
          
By connecting the results in the preceding section, we get $\boldsymbol{y}^{f} = (\boldsymbol{{y}}^{ct, b}, {w}^{k}_{tp}, \boldsymbol{{y}}^{md, f}, {w}^{k}_{et}, \boldsymbol{{y}}^{ce, f})$ as the forward situation and $\boldsymbol{y}^{b}$ means the backward situation. GRU networks are used as encoders to process sequences in different situations, which do not share parameters. And the direction is predicted by:
{\setlength\abovedisplayskip{1em}
\setlength\belowdisplayskip{1em}
\begin{align}
	p(d | \boldsymbol{{y}}^{f}, \boldsymbol{{y}}^{b}) &= sigmoid(\mathbf{{W}}^{d}[\tilde{h}^{d, f}, \tilde{h}^{d, b}]) \\
	\tilde{h}^{d, *} &= \sum_{i = 1}^{T'} \mathbf{GRU}(y_i^{*})
\end{align}}where ${*} \in$ \{${f}$, ${b}$\} means forward or backward.
After the operation completes, one of the sequences $\boldsymbol{y}^{f}$ and $\boldsymbol{y}^{b}$ should conform to our expectations. 

\section{Experiment}
\label{sec:ourExper}

\subsection{Data} 

We evaluated and trained E-SCBA on the emotional conversation dataset NLPCC2017.
There are a total of 1,119,201 Chinese post-reply pairs in the set. 
The dictionaries mentioned in Section \ref{sec:wordPredictor} were used to mark the conversation. The cases whose replies contain both emotion keywords and topic keywords account for 42.6\% (476,121) of the total\footnote{Please note that we did not use the original labels of the dataset, but the emotion categories of the keywords as labels to avoid unnecessary bias. For cases that contain multiple topic keywords or emotion keywords, we chose the keywords that appear less frequently to reduce imbalances.}, which are suitable data for the bidirectional-asynchronous decoder. We randomly sampled 8,000 for validation, 3,000 for testing and the rest for training.
We also sampled another 60,000 pairs from the training set to train the LDA model\footnote{High frequency words and stop words, which have no benefit to the topics, were removed in advance.} mentioned in Section \ref{sec:wordPredictor}.
Besides, an error analysis is presented based on a Chinese movie subtitle dataset which is collected from the Internet. 

\begin{table*}[tp]
\centering
	\subtable{
	\centering
	\normalsize
	\setlength{\tabcolsep}{1.2mm}{
	\begin{tabular}{l||c|c|c|c|c|c|c|c|c|c|c|c}
		\hline
		\multirow{2}{*}{\small{Method}} & \multicolumn{3}{c|}{\bf{\small{Overall}}} & \multicolumn{3}{c|}{\small{Happy}} & \multicolumn{3}{c|}{\small{Like}} & \multicolumn{3}{c}{\small{Surprise}}\\
		\cline{2-4} \cline{5-7} \cline{8-10} \cline{11-13}
		&\small{C}&\small{L}&\small{E}&\small{C}&\small{L}&\small{E}&\small{C}&\small{L}&\small{E}&\small{C}&\small{L}&\small{E}\\
		\hline
		\small{S2S}			& \small{1.301} & \small{0.776} & \small{0.197}		& \small{1.368} & \small{0.924} & \small{0.285} 		& \small{1.341} & \small{0.757} & \small{0.217} 		& \small{1.186} & \small{0.723} & \small{0.076}\\
   		\small{S2S-AW}	  		& \small{1.348} & \small{1.063} & \small{0.231} 		& \small{1.437} & \small{1.097} & \small{0.237} 		& \small{1.418} & \small{1.125} & \small{0.276}		& \bf{\small{1.213}} & \bf{\small{0.916}} & \small{0.105}\\
    		\small{E-SCBA}	& \bf{\small{1.375}} & \bf{\small{1.123}} & \bf{\small{0.476}} 	& \bf{\small{1.476}} & \bf{\small{1.286}} & \bf{\small{0.615}} 	& \bf{\small{1.437}} & \bf{\small{1.173}} & \bf{\small{0.545}} 	& \small{1.197} & \small{0.902} & \bf{\small{0.245}}\\  
    		\hline
		\multirow{2}{*}{\small{Method}} & \multicolumn{3}{c|}{\small{Sad}} & \multicolumn{3}{c|}{\small{Fear}} & \multicolumn{3}{c|}{\small{Angry}} & \multicolumn{3}{c}{\small{Disgust}}\\
		\cline{2-4} \cline{5-7} \cline{8-10} \cline{11-13}
		&\small{C}&\small{L}&\small{E}&\small{C}&\small{L}&\small{E}&\small{C}&\small{L}&\small{E}&\small{C}&\small{L}&\small{E}\\
		\hline
		\small{S2S}		& \small{1.393} & \small{0.928} & \small{0.237}		& \small{1.245} & \small{0.782} & \small{0.215} 		& \bf{\small{1.205}} & \small{0.535} & \small{0.113}	& \small{1.368} & \small{0.680} & \small{0.236} \\
   		\small{S2S-AW}	  	& \small{1.423} & \small{1.196} & \small{0.293} 		& \small{1.260} & \small{1.105} & \small{0.272} 		& \small{1.198} & \bf{\small{0.860}} & \small{0.182}	& \small{1.488} & \small{1.145} & \small{0.253} \\
    		\small{E-SCBA} 	& \bf{\small{1.497}} & \bf{\small{1.268}} & \bf{\small{0.525}} 	& \bf{\small{1.268}} & \bf{\small{1.124}} & \bf{\small{0.453}} 	& \small{1.110} & \small{0.822} & \bf{\small{0.347}}	& \bf{\small{1.637}} & \bf{\small{1.289}} & \bf{\small{0.603}} \\  
    		\hline
	\end{tabular}}}  
	\caption{The results of human annotations (C = Consistency, L = Logic, E = Emotion).} 
	\label{table:cle}
\end{table*}

\begin{table}[t]
\newcommand{\tabincell}[2]{\begin{tabular}{@{}#1@{}}#2\end{tabular}}
	\centering
	\normalsize
	\setlength{\tabcolsep}{1.4mm}{
	\begin{tabular}{l||c|c|c|c|c}
		\hline
		\small{Method}  & \tabincell{c}{\small{G--M}} & \tabincell{c}{\small{E--A}} & \tabincell{c}{\small{V--E}} & \tabincell{c}{\small{D--1}} & \tabincell{c}{\small{D--2}}\\
		\hline
		\small{S2S} & \small{0.297} & \small{0.382} & \small{0.284} & \small{0.086} & \small{0.212}\\
		\small{S2S-STW} & \small{0.328} & \small{0.433} & \small{0.327} & \small{0.135} & \small{0.343}\\
		\small{S2S-SEW} & \small{0.322} & \small{0.421} & \small{0.319} & \small{0.146} & \small{0.364}\\
		\small{S2S-AW} & \small{0.363} & \small{0.485} & \small{0.352} & \small{0.162} & \small{0.417}\\
		\small{E-SCBA} & \bf{\small{0.405}} & \bf{\small{0.553}} & \bf{\small{0.395}} & \bf{\small{0.218}} & \bf{\small{0.582}}\\
		\hline
	\end{tabular}}
	\caption{The results of automatic evaluation (G--E = Greedy Matching, E--A = Embedding Average, V--E = Vector Extrema).}
	\label{table: ED}
\end{table}

\subsection{Metrics}

To evaluate our approach, we use the metrics as below:

{\bfseries Embedding-based Metrics:} We measure the similarity computed by cosine distance between a candidate reply and the target reply using sentence-level embedding, 
following the work in \cite{liu2016not, serban2017hierarchical}.

{\bfseries Distinct Metrics:} By computing the number of different unigrams (Distinct-1) and bigrams (Distinct-2), we measure information and diversity in the candidate replies, following the work in \cite{li2016diversity, xing2017topic}.

{\bfseries Human Annotations:} We asked four annotators to evaluate the replies\footnote{700 conservations in total, 100 for each emotion category, were sampled randomly from the test set.} 
generated from our approach and baselines from \emph{Consistency}, \emph{Logic} and \emph{Emotion}.
\emph{Consistency} measures fluency and grammaticality of the reply on a three-point scale: 0, 1, 2; 
\emph{Logic} measures the degree to which the post and the reply logically match on a three-point scale\footnote{If a reply is too short or turns up frequently, it would be annotated as either 0 or 1 (if the annotator thought the reply related to the post), like ''Me too'' and ''I think so''.} as above; 
\emph{Emotion} judges whether the reply includes the right emotion. A score of 0 means the emotion is wrong or there is no emotion, and a score of 1 is the opposite.

\subsection{Baselines}

In the experiments, E-SCBA is compared with the following baselines:

{\bfseries S2S}: the general seq2seq model with attention method \cite{bahdanau2014neural}.

{\bfseries S2S-STW}: the model uses a synchronous method that starts generating its reply solely and directly from the topic keyword.

{\bfseries S2S-SEW}: the model uses a synchronous method that starts generating its reply solely and directly from the emotion keyword.

{\bfseries S2S-AW}: the model uses an asynchronous method the same as \cite{mou2016sequence}.

The synchronous method in S2S-STW and S2S-SEW was mentioned in \cite{mou2015backward}, acting as the contrast to the asynchronous models. 

\begin{CJK*}{UTF8}{gbsn}
\begin{table*}[t]
\newcommand{\tabincell}[2]{\begin{tabular}{@{}#1@{}}#2\end{tabular}}
	\centering
	\small
	\setlength{\tabcolsep}{2.0mm}{
	\begin{spacing}{1.19}
	\begin{tabular}{|c|c|c|c|c|c|}
		\hline
		\bf{Post} & \bf{Emotion} & \bf{Chinese} & \bf{English(translated)}\\
		\hline
		{\tabincell{c}{受当局追捕，我们只能秘密活动。\\Hunted by the authorities, \\we work in secret.}} & Disgust & 有一种被{\color{red} \bf{嘲讽}}的{\color{blue} \bf{感觉}}。 & There is a  {\color{blue} \bf{sense}} of being {\color{red} \bf{mocked}}.\\
		\hline 
		{\tabincell{c}{我一直在观察你们的制作过程。\\I have been observing your \\production process.}} & Happy & 我{\color{red} \bf{希望}}你继续{\color{blue} \bf{努力}} & I {\color{red} \bf{hope}} you keep {\color{blue} \bf{working hard}}\\
		\hline
		{\tabincell{c}{假如你那天阻止了他，事情\\就不会这样了。\\If you had stopped him that day,\\ things would have been different.}} & Sad & 不要{\color{red} \bf{回忆}}过去的{\color{blue} \bf{日子}}了！ & Don't {\color{red} \bf{look back}} on the {\color{blue} \bf{past}}!\\
		\hline
		{\tabincell{c}{普通民众中也可以拥有某些\\权利的概念。\\The general public can also have \\some concepts of rights.}} & Like & 可以{\color{red} \bf{作为}}他们{\color{blue} \bf{生活}}的一部分 & This can {\color{red} \bf{be}} a part of their {\color{blue} \bf{life}}\\
		\hline
		{\tabincell{c}{我们现在也可以离开这里了。\\We can also leave here now.}} & Sad & {\color{blue} \bf{挺}}{\color{red} \bf{遗憾}}的。 & What a {\color{red} \bf{pity}}.\\
		\hline
		{\tabincell{c}{那个人最期待的是像你这样的\\美女送这个箱子。\\The man is looking forward to having a \\beautiful girl like you to send this box.}} & Disgust & 我{\color{blue} \bf{期待}}的是像你这样的{\color{red} \bf{流氓}} & {\tabincell{c}{I am {\color{blue} \bf{looking forward to}} a \\{\color{red} \bf{hooligan}} like you}}\\
		\hline
	\end{tabular}
	\end{spacing}}
	\caption{Sampled conversations with a corresponding emotion from the Chinese movie subtitle data.}
	\label{table: MovieCase}
\end{table*}
\end{CJK*}

\subsection{Results and Discussion}

The results of automatic evaluation are shown in Table \ref{table: ED}.
Compared with the best model (S2S-AW) that considers only a single factor, E-SCBA makes significant improvement on the distinct metrics (+0.056 and +0.165), which verifies the effectiveness of taking both emotion and topic information into account to improve the diversity.
Likewise, our approach also respectively achieves 0.042, 0.068 and 0.043 gains on G-M, E-A and V-E, benefiting from the compound information that captures the thrust of human conversation so that E-SCBA has a better ability to learn the goal distribution.
Furthermore, the grades of the asynchronous models are higher than the synchronous models on both kinds of metrics, showing that the asynchronous method is a more suitable way for content-introducing conversation generation.

Table \ref{table:cle} depicts the human annotations (\emph{t-test}: $p$ $<$ 0.05 for \emph{C} and \emph{L}, $p$ $<$ 0.01 for \emph{E}). 
Overall, E-SCBA outperforms S2S-AW on all three metrics, where the compound information plays a positive role in the comprehensive promotion.
However, in \emph{Surprise} and \emph{Angry}, the grades of \emph{Consistency} and \emph{Logic} are not satisfactory, since the data for them are much less than others (\emph{Surprise} (1.2\%) and \emph{Angry} (0.7\%)).
Besides, the score of \emph{Emotion} in \emph{Surprise} has a big difference from others. We think the reason is that the characteristic of \emph{Surprise} overlaps with other categories that have much more data, such as \emph{Happy}, which interferes with the learning efficiency of the approach in \emph{Surprise}. Meanwhile, it is harder for annotators to determine which one is the right emotion.

\subsection{Case Study and Error Analysis}
In this section, we sampled some typical cases from a Chinese movie subtitle dataset to do a further error analysis. The cases are shown in Table \ref{table: MovieCase}. 
The post of weibo and movie subtitle are applied in different scenes to obey different distributions. The weaker correlation between training sets and test sets can present a more reliable study.

The first three conversations are positive samples and others are negative samples that have content with flaws. 
For the reply in the antepenultimate line, its problem is the faint emotion. Since the emotion keyword in this sentence is a polysemic word, and it expresses a meaning with no emotion here. Under diverse circumstances, a polysemic word probably have different meanings, emotional or neutral. For example, the word ''like'' can be a generic word when it denotes \emph{similar}, but it can also be an emotion word when it denotes \emph{enjoy}. Same situation also occurs in Chinese.
Besides, we notice that if the LDA model pick a meaningless topic keyword from the dictionary, our approach may have a difficulty in generating a diverse and long reply, as the reply in the penultimate line. The lack of information causes generic replies which are consisted of few words generated from the networks.
The last line presents another limitation. The emotion keyword \emph{hooligan} corresponds to the post and the topic keyword \emph{looking forward to} is meaningful, but the combination of them, \emph{looking forward to} a \emph{hooligan}, does not conform to the normal logic.
This situation is caused by the fact that two kinds of keywords are generated independently before decoding, and it may cause a mismatch.
In the future, we will try to explore different network architectures to make keywords interact with each other during the generation.

\section{Conclusion}
\label{sec:Conclusion}
In this paper, we proposed a novel conversation generation approach (E-SCBA) to make a more comprehensive optimization for the quality of reply, which introduces both emotion and topic knowledge into the generation.
The newly designed decoder makes use of syntactic knowledge to constrain generation and ensures fluency and grammaticality of reply.
Experiments show that our approach can generate replies that have rich diversity and feature both emotion and logic.

\bibliography{emnlp2018}
\bibliographystyle{acl_natbib_nourl}

\end{document}